\documentclass{article}

\usepackage{arxiv}

\usepackage[utf8]{inputenc} % allow utf-8 input
\usepackage[T1]{fontenc}    % use 8-bit T1 fonts
\usepackage{hyperref}       % hyperlinks
\usepackage{url}            % simple URL typesetting
\usepackage{booktabs}       % professional-quality tables
\usepackage{amsfonts}       % blackboard math symbols
\usepackage{nicefrac}       % compact symbols for 1/2, etc.
\usepackage{microtype}      % microtypography
\usepackage{lipsum}		% Can be removed after putting your text content
\usepackage{graphicx}
\usepackage{doi}
\usepackage{natbib}
\usepackage{amsthm}
\usepackage{amsmath}
\usepackage{multirow}
\usepackage{multicol}
\usepackage{color}
\usepackage{xcolor}
\usepackage{colortbl}
\usepackage{bbding}
\definecolor{C1}{HTML}{93BFCF}
\definecolor{C2}{HTML}{A0C3D2}
\definecolor{C3}{HTML}{BDCDD6}
\definecolor{C4}{HTML}{EEE9DA}
\definecolor{C5}{HTML}{FFF1DC}
\definecolor{C6}{HTML}{E8D5C4}
\definecolor{C7}{HTML}{EEEEEE}
\definecolor{C8}{HTML}{BCEE68}

\title{RefGaussian: Disentangling Reflections from 3D Gaussian Splatting for Realistic Rendering}

\author{Rui Zhang$^{*,1}$, Tianyue Luo$^{*,1}$, Weidong Yang{$^{\dagger,1}$}, Ben Fei$^{\dagger,1}$, \\
\textbf{Jingyi Xu}$^{1}$, \textbf{Qingyuan Zhou}$^{1}$, \textbf{Keyi Liu}$^{1}$, \textbf{Ying He}{$^{\ddagger,2}$}\\
	$^1$Fudan University, $^2$Nanyang Technological University\\
    \texttt{22210240379@m.fudan.edu.cn}, 
    \texttt{bfei21@m.fudan.edu.cn}, \texttt{wdyang@fudan.edu.cn},
    \texttt{yhe@ntu.edu.sg}\\
}

\date{}

\renewcommand{\shorttitle}{\textit{} RefGaussian}

\begin{document}
\maketitle

\begin{abstract}
	3D Gaussian Splatting (3D-GS) has made a notable advancement in the field of neural rendering, 3D scene reconstruction, and novel view synthesis. Nevertheless, 3D-GS encounters the main challenge when it comes to accurately representing physical reflections, especially in the case of total reflection and semi-reflection that are commonly found in real-world scenes. This limitation causes reflections to be mistakenly treated as independent elements with physical presence, leading to imprecise reconstructions. Herein, to tackle this challenge, we propose RefGaussian to disentangle reflections from 3D-GS for realistically modeling reflections. Specifically, we propose to split a scene into transmitted and reflected components and represent these components using two Spherical Harmonics (SH). Given that this decomposition is not fully determined, we employ local regularization techniques to ensure local smoothness for both the transmitted and reflected components, thereby achieving more plausible decomposition outcomes than 3D-GS. Experimental results demonstrate that our approach achieves superior novel view synthesis and accurate depth estimation outcomes. Furthermore, it enables the utilization of scene editing applications, ensuring both high-quality results and physical coherence.
\end{abstract}

% keywords can be removed

\newcommand\blfootnote[1]{%
\begingroup
\renewcommand\thefootnote{}\footnote{#1}%
\addtocounter{footnote}{-1}%
\endgroup
}
\blfootnote{{$*$}Equal contribution, {$\dagger$}Corresponding author, {$\ddagger$ Project leader}.}

\section{Introduction}
\label{sec:Introduction}
Novel view synthesis from multi-view images has been recognized as a pivotal research area in computer graphics and vision for its potential capability in applications like virtual reality, augmented reality, and 3D reconstruction. 
As a pioneering work, NeRF (Neural Radiance Fields)~\citep{reiser2021kilonerf}, an implicit scene representation with neural radiance fields, represents a significant breakthrough in novel view synthesis with photo-realistic view-dependent rendering capabilities.
NeRF utilizes a fully connected deep neural network to map spatial coordinates and viewing directions to the corresponding colors and volume densities, which are later accumulated by the volumetric rendering to form the final color of the pixels. 
Despite its remarkable performance, NeRF and its variants~\citep{deng2022depth,barron2021mip,pumarola2021d} are constrained by computational complexity and the time-consuming optimization process. 
On the contrary, the newly proposed 3D Gaussian Splatting (3D-GS)~\citep{kerbl20233d} has emerged as a promising alternative approach, which represents scenes explicitly with anisotropic 3D Gaussians.
%each Gaussian ball is defined by the mean position, covariance matrix for the shape control, opacity and  spherical harmonics parameters for modeling view-dependent color.
3D-GS first initializes Gaussians parameters from the Structure-from-motion (SfM)~\citep{schonberger2016structure} point cloud. During the rendering process, these Gaussians are splatted to the 2D image plane, and their colors are computed from the Spherical Harmonics (SH) parameters. Then, point-based differentiable rasterization is used to render the image from novel viewpoints. 
With efficient optimization and rendering, 3D-GS outperforms NeRF in real-time applications while maintaining a comparative rendering quality.

\begin{figure}[ht]
 \includegraphics[width=0.9\linewidth]{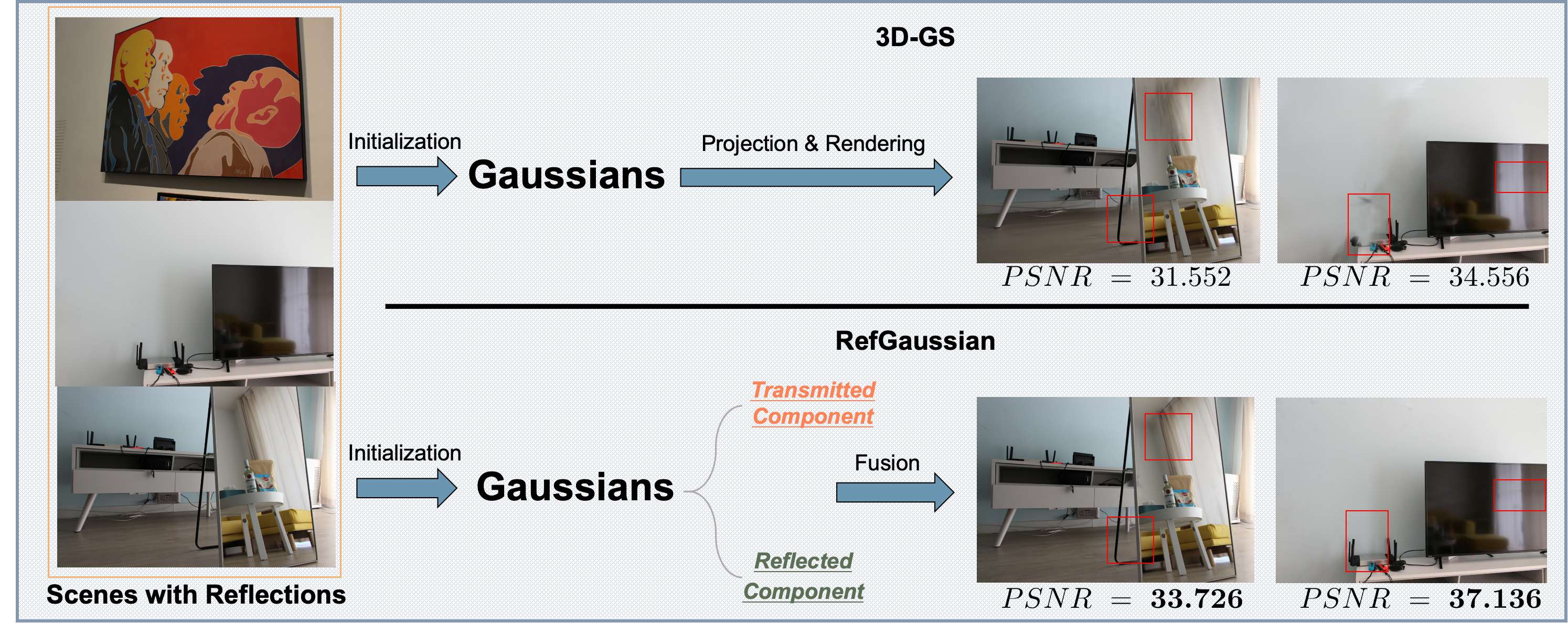}
 \centering
  \caption{\textbf{Novel view synthesis with RefGaussian, fulfilling better reflections modeling than original 3D-GS.}}
\label{fig:teaser}
\end{figure}

Nonetheless, both NeRF and 3D-GS fall short of accurately reconstructing scenes that feature reflective surfaces such as mirrors, glass, and screens (Figure~\ref{fig:teaser}). This limitation arises from the fact that the impact of reflections is disregarded in both of these representations.
As a result, these models often represent reflected contents as physically existing entities, which significantly undermines the fidelity and coherence of the rendered scenes. Moreover, the reflected contents can vary depending on the viewpoint, leading to inconsistencies across multiple views. This multi-view inconsistency can further introduce conflicts during optimization and result in blurred outcomes.

Several methods are proposed to tackle this problem of scene reconstruction with reflections. Mirror-NeRF~\citep{zeng2023mirror} additionally parameterizes the spatial points with the reflection probability and traces the rays following Whitted Ray Tracing~\citep{whitted2005improved}. 
The pixel color is blended by the color of the ray and its reflected ray according to the volume-rendered reflection probability. 
TraM-NeRF~\citep{vanholland2023tramnerf} models the mirror reflections explicitly using physically plausible materials and integrates the volume rendering formulation with Monte-Carlo methods to estimate the reflected radiance. 
Approaches such as NeRFReN~\citep{guo2022nerfren} and MS-NeRF~\citep{yin2023multi} are capable of modeling more diverse reflections into distinct components or subspaces via partitioning scenes.
However, these methods inevitably inherit the limitation of slow optimization as an unresolved challenge (Table.~\ref{tab:method_com}).
%However, these methods still encounter the challenge of slow optimization as an inherent limitation.
The Mirror-3DGS~\citep{meng2024mirror} and MirrorGaussian~\citep{liu2024mirrorgaussian} methods, which are based on 3D-GS, utilize the plane equation to represent the mirror and exploit the symmetry to combine the real-world image with its mirrored counterpart for the ultimate outcome. Nonetheless, these approaches necessitate accurate mirror masks to assist in the mirror parameterization and solely focus on modeling specular reflections (Table.~\ref{tab:method_com}).

Therefore, effectively synthesizing scenes with general reflections based on 3D-GS is still an underexplored area.
In this study, we introduce RefGaussian, a novel disentangled mask-free method that enhances the fidelity of 3D-GS in accurately representing physical reflections.
In practice, we propose to decompose the rendering process into two distinct components: the transmitted component and the reflected component.
Instead of initializing two sets of Gaussians for the modeling of each part, we present a joint training and synchronous rendering pipeline by extending the Gaussian representation with three extra reflection-related parameters, which are reflection Spherical Harmonics, reflection opacity, and reflection confidence.
This design helps to relieve the multi-view inconsistency and stabilize the training since it forces the reflected contents to under the control of Gaussians to which they virtually belong, rather than treated as separate physical entities.
%be controlled by the Gaussians they virtually belong to, rather than treated separately as physical entities. 
During the training, the Gaussians are first projected to the 2D image plane and undertake a synchronous rendering process with two sets of Spherical Harmonics coefficients to generate the transmitted image and the reflected image. 
Meanwhile, the reflection map is generated by accumulating reflection confidences in a similar way to the alpha blending of color. The final image is produced by fusing the transmitted image with the reflection map weighted reflected image. 

However, it is challenging to separate the reflections from the scene without the guidance of a mask. An incomplete decomposition of the two components could lead to conflicts during training and a drastic performance drop in the model. Consequently, the direct outcome is that the reflected contents degenerate into translucent fogs and blurs at the pixel level. 
Nevertheless, the depth map of the rendered image is also compromised.
NeRFReN~\citep{guo2022nerfren} introduces a depth smoothness loss to effectively regularize the depth map of the transmitted component and mitigates the information leakage between components. However, this constraint primarily addresses abrupt depth changes and does not adequately alleviate conflicts in color-related parameters.
Drawing inspiration from the classic bilateral filter, which utilizes both domain and range filters to achieve image smoothing, we propose a bilateral smoothness constraint. This constraint aims to regulate the decomposition process by taking into account both depth and color information. In addition to the constraint imposed by the holistic view, a smoothness loss is also applied to the reflection map to ensure coherence in the reflected component.
With the hybrid design of the scene decomposition and bilateral smoothness regularization, RefGaussian is able to reconstruct the scene with reflections both efficiently and faithfully. 
Furthermore, by utilizing the acquired reflection map, which represents the likelihood of a pixel containing reflections, we can perform various reflection manipulations such as highlighting and shading.

Our contributions can be concluded as follows:

1) We propose RefGaussian to model general reflections by decomposing the scene into the transmitted component and the reflected component. The framework adopts a joint training and synchronous rendering scheme with no need for additional Gaussian balls and achieves efficient and realistic rendering. 

2) We introduce the bilateral smoothness and the reflection map smoothness constraints to facilitate the scene decomposition. Both depth variations and color differences are considered and then optimized in a collaborative manner. 

3) RefGaussian demonstrates superior performance in view synthesis using real-captured datasets that include reflections. Additionally, it achieves competitive results in more general scenarios. Furthermore, we also highlight the potential of RefGaussian in scene editing tasks, specifically in the manipulation of reflections.

\section{Related Works}
\label{sec:Related}

\subsection{Novel View Synthesis}
Novel View Synthesis (NVS) aims to generate photo-realistic images from new viewpoints using a set of calibrated images capturing a 3D scene~\citep{wang2023sparsenerf,chan2023generative}. NeRF~\citep{mildenhall2021nerf} is a notable breakthrough in this field, employing MLPs (Multilayer Perceptrons) to estimate density and view-dependent colors for densely sampled points through ray tracing. 
It then utilizes volume rendering to generate novel views. However, NeRF suffers from computational demands due to intensive point sampling and expensive MLP queries, resulting in inefficient optimization and slow rendering~\citep{xu20223d}. 
This limitation makes NeRF unsuitable for interactive applications. Achieving a balance between quality and efficiency remains challenging for NeRF-based methods~\citep{muller2022instant,reiser2021kilonerf,reiser2023merf}.
Concurrently, point-based rasterization methods have shown impressive results by offering a favorable combination of rendering quality and computational efficiency~\citep{kopanas2021point,ruckert2022adop}. One noteworthy example is 3D-GS~\citep{kerbl20233d,fei20243d}, which has gained significant attention. 3D-GS explicitly represents a 3D scene using anisotropic 3D Gaussian points. Each point contains opacity and a set of Spherical Harmonic coefficients to model view-dependent colors. By employing a hardware-accelerated point-based rasterizer instead of computationally intensive ray tracing, 3D-GS achieves both high-quality and real-time rendering simultaneously. Recent efforts have focused on further enhancing the quality of 3D-GS. For instance, Scaffold-GS~\citep{lu2023scaffold} introduces an anchor-based approach to distribute 3D Gaussians, resulting in higher rendering quality. Mip-Splatting~\citep{yu2023mip} addresses the aliasing problem in 3D-GS by implementing a 3D smoothing filter and a 2D Mip filter, leading to alias-free results. 
However, both NeRF-based and 3D-GS-based methods still face challenges when it comes to reconstructing scenes with intricate reflections, which can vary significantly from different viewpoints. Existing methods struggle to model such variations using MLPs or single SH functions.

\begin{figure*}[t]
    \centering
    \includegraphics[width=\linewidth]{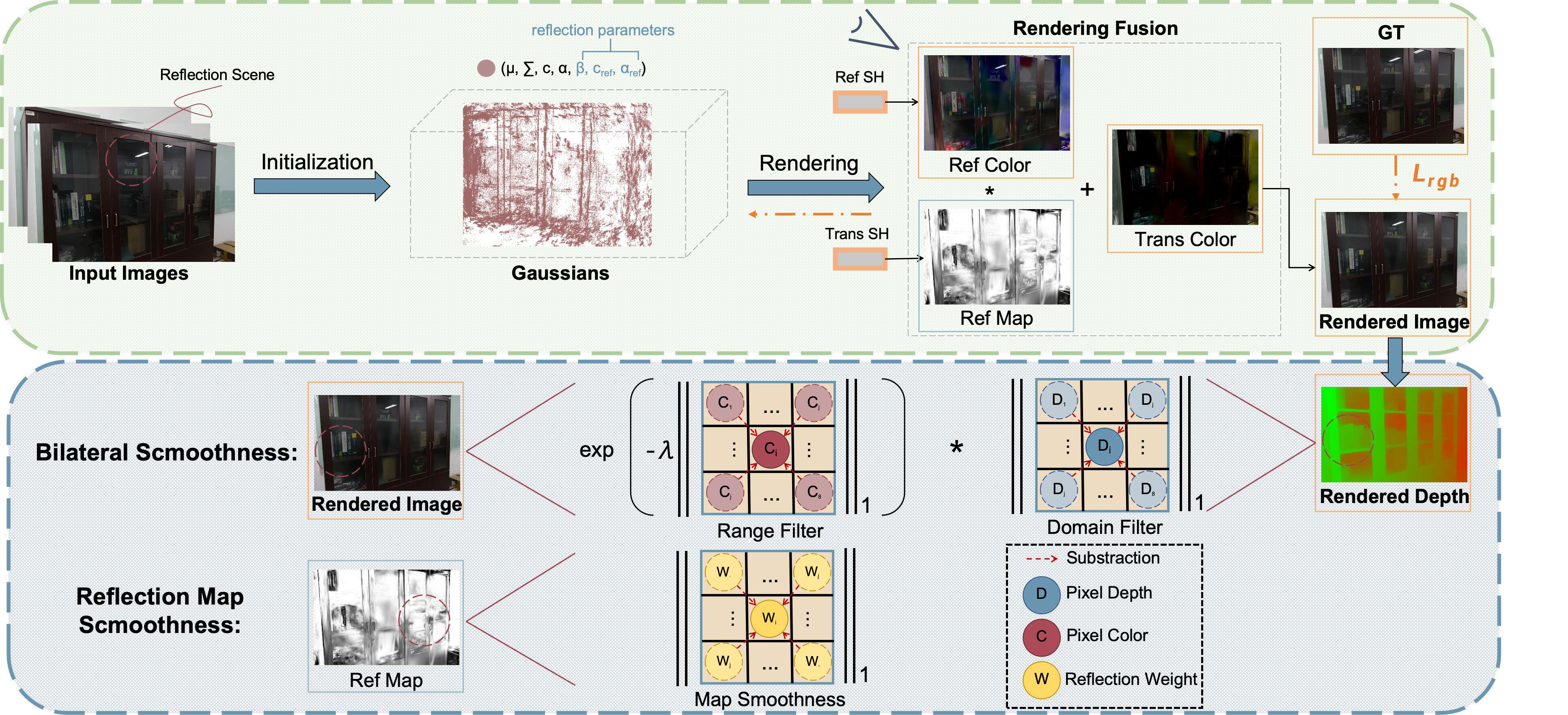}
    \caption{The overview of our proposed RefGaussian. The RefGaussian framework is specifically designed to accurately model general reflections within a scene by effectively decomposing it into two distinct components: the transmitted component and the reflected component. RefGaussian eliminates the requirement for additional 3D Gaussians, resulting in a more efficient and realistic rendering process. By incorporating bilateral smoothness and reflection map smoothness, the framework enables effective scene decomposition. Moreover, both depth variations and color differences are taken into account and collaboratively optimized to further enhance the overall rendering quality.}
    \label{fig:pipeline}
\end{figure*}

\subsection{Reflection Modeling in NeRF}

Reflections are a prevalent phenomenon in the real world, but accurately modeling them can be difficult without employing techniques such as ray tracing to simulate light paths.
In computer graphics, the Screen Space Reflection technique has been developed to simulate reflection effects at a minimal computational expense. However, synthesizing novel views has long posed significant challenges, yet these areas have received limited exploration~\citep{xu2021scalable,sinha2012image}.
RefNeRF~\citep{verbin2022ref} effectively separates light into two components: diffuse and specular. It achieves this by utilizing a reflection-dependent radiance field to represent reflections considering the viewing direction accurately. 
On the other hand, SpecNeRF~\citep{ma2023specnerf} introduces a Gaussian directional encoding that is adaptable through learning. This enhancement allows for a more accurate representation of view-dependent effects, particularly close to the light source.
Taking a step further, Guo et al.~\citep{guo2022nerfren} introduced NeRFReN, an extension of NeRF designed to represent scenes with reflections accurately. The proposed approach involves dividing a scene into two components: the transmitted and reflected parts. Each component is then modeled using separate neural radiance fields. Given the inherent challenges of decomposing a scene in this manner, the authors leverage geometric priors and employ meticulously designed training strategies to attain plausible decomposition outcomes.
However, the persistently slow rendering speed poses a significant obstacle to the broader applicability of these NeRF-based methods (Table.~\ref{tab:method_com}). The approach proposed in this paper is based on reflection decoupling and innovatively applied to the modeling of reflective scenes within the 3D-GS framework. This strategy maintains the fast training and real-time rendering advantages of 3D-GS.

\subsection{Reflection Modeling in 3D Gaussian Splatting}
Encouraged by the speed and rendering quality of 3D-GS, several recent works tend to model the reflection in 3D-GS~\citep{jiang2023gaussianshader,gao2023relightable,malarzgaussian}.
GaussianShader~\citep{jiang2023gaussianshader} applies a simplified shading function to 3D-GS in order to enhance neural rendering in scenes featuring reflective surfaces, while also ensuring efficient training and rendering. 
Relightable 3D Gaussian~\citep{gao2023relightable} assigns normal, BRDF properties, and incident light information to each 3D Gaussian point, allowing for the modeling of per-point light reflectance. 
VDGS~\citep{malarzgaussian} combines the strengths of NeRF and 3D-GS, resulting in swift training and inference, as well as effective modeling of shadows, light reflections, and transparency to generate realistic 3D objects. 
However, it is important to note that these methods are currently limited to single objects rather than entire scenes (Table.~\ref{tab:method_com}).
MirrorGaussian~\citep{liu2024mirrorgaussian} accomplishes both high-quality reconstruction and real-time rendering for scenes that feature mirrors. The approach introduces an explicit point-cloud-based representation, leveraging the mirror symmetry existing between the physical space and the virtual mirror space.
Concurrently, Mirror-3DGS~\citep{meng2024mirror} tackles the issue of capturing and rendering reflections in scenes that include mirrors. To achieve this, Mirror-3DGS utilizes a two-stage training process. In the first stage, it filters mirror Gaussians, and in the second stage, it uses this filtered information to precisely estimate the mirror plane and derive the parameters for the mirrored camera.
Nevertheless, the practical utilization of Mirror-3DGS and MirrorGaussian is limited due to the requirement of a mask to identify the location of the mirror, which is rarely available in real-world scenarios (Table.~\ref{tab:method_com}). 
Therefore, in this paper, we propose RefGaussian aiming to model complex reflections in real-world scenes without assigning manually labeled masks.

\begin{table*}[ht]
\centering
\caption{\textbf{Comparison of different NeRF-based and 3D-GS derived methods for reflection modeling.}}
\resizebox{\linewidth}{!}{
\begin{tabular}{l|ccc}
\toprule[1.5pt]
                        & Real-world Scene Rendering & Hard Mask Free & Fast Rendering Speed \\ \midrule[1pt]
NeRF-D~\citep{guo2022nerfren}                  & \CheckmarkBold                & \XSolidBrush              & \XSolidBrush                    \\
RefNeRF~\citep{verbin2022ref}                 & \CheckmarkBold                & \CheckmarkBold              & \XSolidBrush                    \\
NeRFReN~\citep{guo2022nerfren}                 & \CheckmarkBold                & \XSolidBrush              & \XSolidBrush                    \\
GaussianShader~\citep{jiang2023gaussianshader}          & \XSolidBrush               & \XSolidBrush             & \CheckmarkBold                    \\
Relightable 3D Gaussian~\citep{gao2023relightable} & \XSolidBrush                & \XSolidBrush              & \CheckmarkBold                   \\
MirrorGaussian~\citep{liu2024mirrorgaussian}          & \CheckmarkBold                & \XSolidBrush              & \CheckmarkBold                    \\
Mirror-3DGS~\citep{meng2024mirror}             &  \CheckmarkBold                & \XSolidBrush              & \CheckmarkBold                    \\ \midrule[1pt]
RefGaussian             & \CheckmarkBold                & \CheckmarkBold              & \CheckmarkBold                    \\ \bottomrule[1.5pt]
\end{tabular}
}
\label{tab:method_com}
\end{table*}

\section{RefGaussian}

\label{sec:Methods}

\subsection{Preliminary}

Gaussian Splatting Radiance Field~\citep{kerbl20233d,fei20243d} is an explicit scene representation based on radiance fields. It utilizes a multitude of 3D anisotropic Gaussians to represent the radiance field, with each Gaussian being modeled using a 3D Gaussian distribution (Eq.~\ref{eq:bg1}).
More concretely, each 3D anisotropic Gaussian has mean $\mathcal{M} \in \mathbb{R}^3$, covariance $\Sigma$, opacity $\theta \in \mathbb{R}$ and spherical harmonics parameters $\mathcal{C} \in \mathbb{R}^k$ ( $k$ is the degrees of freedom) for modeling view-dependent color. For regularizing optimization, the covariance matrix is further decomposed into rotation matrix $\mathbf{R}$ and scaling matrix $\mathbf{S}$ by Eq~\ref{eq:bg2}. These matrices are further represented as quaternions $r \in \mathbb{R}^4$ and scaling factor $s \in \mathbb{R}^3$.
% \vspace{-0.1cm}
\begin{equation}
    G(X)=e^{-\frac{1}{2} \mathcal{M}^T \Sigma^{-1} \mathcal{M}}
    \label{eq:bg1}
\end{equation}
\begin{equation}
    \Sigma=\mathbf{R S S}^T \mathbf{R}^T
    \label{eq:bg2}
\end{equation}
% \vspace{-0.1cm}
For this scene representation, view rendering is performed via point splatting~\citep{yifan2019differentiable}.
Specifically, all Gaussians in the scene are first projected onto the $2 \mathrm{D}$ image plane, and their color is computed from spherical harmonic parameters. Then, for every $16 \times 16$ pixel patch of the final image, the projected Gaussians that intersect with the patch are sorted by depth. For every pixel in the patch, its color is computed by alpha compositing the opacity and color of all the Gaussians covering this pixel by depth order, as in Eq.~\ref{eq:bg3}.
% \vspace{-0.1cm}
\begin{equation}
    C=\sum_{i \in N_{\text {cov }}} c_i \alpha_i \prod_{j=1}^{i-1}\left(1-\alpha_j\right)
    \label{eq:bg3}
\end{equation}
% \vspace{-0.1cm}
where, $N_{\text{cov}}$ represents the splats that cover this pixel, $\alpha_i$ represents the opacity of this Gaussian splat multiplied by the density of the projected 2D Gaussian distribution at the location of the pixel, and $c_i$ represents the computed color.

\subsection{RefGaussian with Spherical Harmonics Decomposition}
However, 3D-GS represents the view-dependent color information with only Spherical Harmonics coefficients, neglecting the effect of physical properties of the object in the scene, e.g., surface reflections. 
To more comprehensively model the scenes with reflections, we extend the 3D-GS with three reflection-related parameters, reflection-SH $\mathcal{C}_{ref}$, reflection opacity $\theta_{ref}$, and reflection confidence $\beta$ to construct our RefGaussian representation as follows:
\begin{equation}
    {RefGaussian} = \{ \mathcal{M}, \Sigma, \mathcal{C}, \theta, \mathcal{C}_{ref}, \theta_{ref}, \beta \}
    \label{eq:refgaussian}
\end{equation}
The reflection confidence $\beta$ is a learnable parameter $\in [0, 1]$ that denotes the probability of an individual Gaussian representing the reflection part. 
This compact and effective design allows us to decompose the rendering into two separate components, i.e., the transmitted rendering component and the reflected rendering component, with each focusing on the different aspect of the scene structure. 
The overview of our proposed pipeline is illustrated in Figure~\ref{fig:pipeline}. 

After projecting the Gaussians onto the 2D image plane, we conduct a dual rendering strategy to generate the transmitted color $\widehat{C}_{trans}$ and the reflected color $\widehat{C}_{ref}$ with respective parameters using Eq.~\ref{eq:bg3}. 

The reflection confidence $\beta$ is accumulated in a similar way to the color rendering to produce the 2D reflection map $W$:
\begin{equation}
    W = \sum_{i \in N_{\text {cov }}} \beta_i \alpha_i \prod_{j=1}^{i-1}\left(1-\beta_j\right)
    \label{eq:reflection map}
\end{equation}

Thus, the final pixel color $C$ is computed with the reflection map weighted fusion of each part:
\begin{equation}
    \widehat{C} = \widehat{C}_{trans} + W * \widehat{C}_{ref}
\label{eq.6}
\end{equation}

\begin{table*}[ht]
\centering
\caption{\textbf{View Synthesis Comparison Results on RFFR Dataset.} We test NeRF, NeRF-D, NeRFReN, 3D-GS and RefGaussian in PSNR, SSIM~\citep{wang2003multiscale} and LPIPS~\citep{zhang2018unreasonable}. Our method exhibits an overall superior performance in PSNR and SSIM. %For a comprehensive comparison, we test the performance on the whole RFFR dataset, Room from the Mip-NeRF360 dataset, and Truck from Tanks\&Temples. 
}
\label{tab:main}
% \resizebox{\linewidth}{!}
{
\begin{tabular}{l|cccccc|c}
\toprule[1pt]
\multicolumn{1}{c}{\multirow{2}{*}{Method}}
            & art1      & \multicolumn{1}{c}{art2}        & \multicolumn{1}{c}{art3}  & \multicolumn{1}{c}{bookcase}   & \multicolumn{1}{c}{tv} & \multicolumn{1}{c}{mirror} & \multicolumn{1}{c}{\textbf{Avg.}}\\ \cline{2-8}
\multicolumn{1}{c}{} & \multicolumn{7}{c}{PSNR $\uparrow$}
 \\ \midrule
NeRF~\citep{mildenhall2021nerf}                                     &    32.354        &  \underline{37.331}        & \textbf{36.653}   & 29.391&31.172 &28.487 &    32.565         \\

NeRF-D~\citep{guo2022nerfren}                                     &32.316 &35.133 & 34.831&29.022 &31.170 &28.322 &  31.799      \\

NeRFReN~\citep{guo2022nerfren}                                     &    \textbf{34.112}        &      \textbf{38.184}     &  35.616  & 
  28.621 & 31.505 &  26.378 &     32.404          \\
3D-GS~\citep{kerbl20233d}         &        32.280     &   36.371           &  \underline{36.619}   & 32.491 &  34.556 &  31.552   &  33.978 \\
\midrule
\textbf{Ours}                                   &   \underline{32.801}           &   36.107                                       &    36.406  & \textbf{32.977} & \textbf{37.136} & \textbf{33.726}    &  \textbf{34.856} \\ \midrule
\midrule
\multicolumn{1}{c}{} & \multicolumn{7}{c}{SSIM $\uparrow$}
 \\ \midrule
NeRF~\citep{mildenhall2021nerf}                                     &    0.919        &   0.935        &  0.937  & 0.853&0.926 &0.881 &  0.909 \\
NeRF-D~\citep{guo2022nerfren}                                     &   0.920         &  0.922    &    0.925 &  0.847  & 0.926 & 0.879 &    0.903                  \\
NeRFReN~\citep{guo2022nerfren}                                     &      0.932      &   0.940    &  0.931  &  0.853  & 0.931 &  0.863 &      0.908            \\
3D-GS~\citep{kerbl20233d}         &      \textbf{0.958}       &           \textbf{0.955}   &   \textbf{0.951}  & \underline{0.947} & \textbf{0.972} & \underline{0.949}     & \underline{0.956}    \\
\midrule
\textbf{Ours}                                   &           \underline{0.954}   &     \textbf{0.955}                                     &    \textbf{0.951}  &  \textbf{0.950} &  \textbf{0.972} & \textbf{0.959}    & \textbf{0.957} \\ \midrule
\midrule
\multicolumn{1}{c}{} & \multicolumn{7}{c}{LPIPS $\downarrow$}
 \\ \midrule
NeRF~\citep{mildenhall2021nerf}                                     &      0.081      & \underline{0.196}          & \textbf{0.208}   & 0.197& \underline{0.072}& \textbf{0.108} &   \underline{0.144}           \\
NeRF-D~\citep{guo2022nerfren}                                     &   \underline{0.079}         &   0.221    &  0.239  &   0.213 &0.074 & \underline{0.110} &   0.156              \\
NeRFReN~\citep{guo2022nerfren}                                     &    \textbf{0.067} &     \textbf{0.191}  &   \underline{0.221} &  \underline{0.189}  &  \textbf{0.068} & 0.113 &     \textbf{0.142}            \\
3D-GS~\citep{kerbl20233d}         &     0.140        &   0.208           &    0.223 &  0.204 &  0.165 & 0.183      &  0.187\\
\midrule
\textbf{Ours}                                   &          0.155   &      0.205                                   &   0.227  & \textbf{0.185} & 0.152 & 0.139  &  0.177 \\ \bottomrule[1pt]
\end{tabular}
}
\end{table*}
\begin{table}[ht]
\centering
\caption{\textbf{Rendering Speed Comparison on RFFR Dataset.} Our method exhibits comparable rendering speed with 3D-GS and exceeds NeRF-based models by a large margin.}
\label{tab:fps}
\resizebox{\linewidth}{!}{
{
\begin{tabular}{l|cccccc|c}
\toprule[1pt]
\multicolumn{1}{c}{\multirow{2}{*}{Method}}
            & art1      & \multicolumn{1}{c}{art2}        & \multicolumn{1}{c}{art3}  & \multicolumn{1}{c}{bookcase}   & \multicolumn{1}{c}{tv} & \multicolumn{1}{c}{mirror} & \multicolumn{1}{c}{\textbf{Avg.}}\\ \cline{2-8}
\multicolumn{1}{c}{} & \multicolumn{7}{c}{FPS $\uparrow$}
 \\ \midrule
NeRF~\citep{mildenhall2021nerf}                                     &     0.041      &     0.041   &   0.041 & 0.041&0.040 &0.040 &       0.041    \\

NeRF-D~\citep{guo2022nerfren}                                       &   0.041        &    0.041    &   0.041 & 0.041& 0.041& 0.041&   0.041 \\

NeRFReN~\citep{guo2022nerfren}                                      &   0.017        &   0.016     &   0.017 &  0.017 & 0.016 &  0.016 &    0.017  \\
3D-GS~\citep{kerbl20233d}         &   \textbf{147.808}        &    \textbf{137.831}    &  \textbf{148.320}  & \textbf{130.586} & \textbf{114.875} & \textbf{130.271} & \textbf{134.949} \\
\midrule
\textbf{Ours}           &    \underline{69.489}       &    \underline{96.041}    &   \underline{101.371}  & \underline{80.802} & \underline{76.482} & \underline{66.952} &  \underline{81.856} \\
\bottomrule[1pt]
\end{tabular}
}
}
\end{table}

\begin{figure*}[ht]
    \centering
    \includegraphics[width=\linewidth]{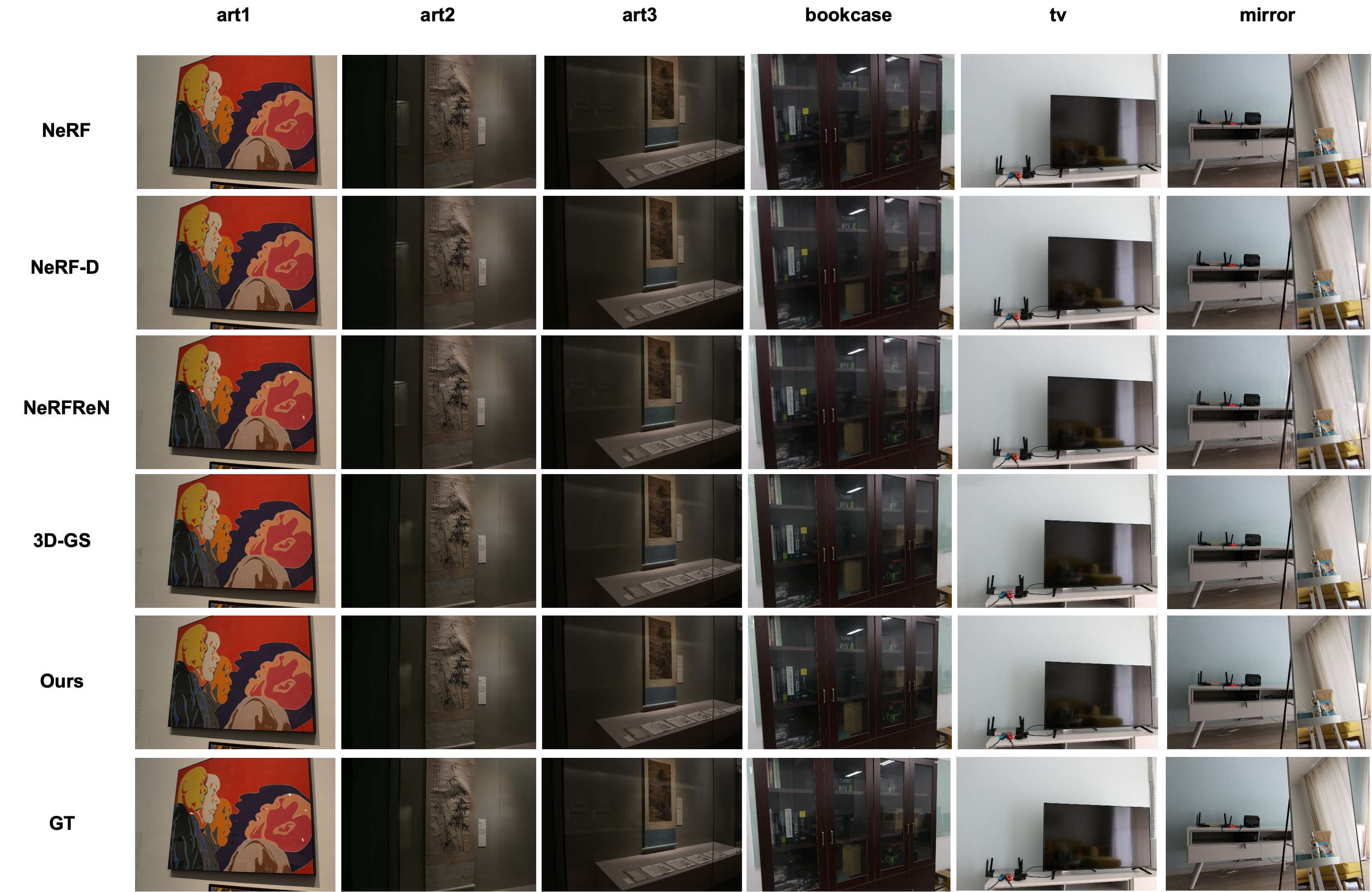}
    \caption{\textbf{Visual comparisons between NeRF, NeRF-D, NeRFReN, 3D-GS and our method.} Our method presents a more detailed and realistic rendering than 3D-GS in all cases. Compared to NeRF-based methods, our methods show comparable results in scenes with semi-reflections and exceed them in scenes containing specular reflections.}
    \label{fig:main}
\end{figure*}

\begin{figure*}[t]
    \centering
    \includegraphics[width=\linewidth]{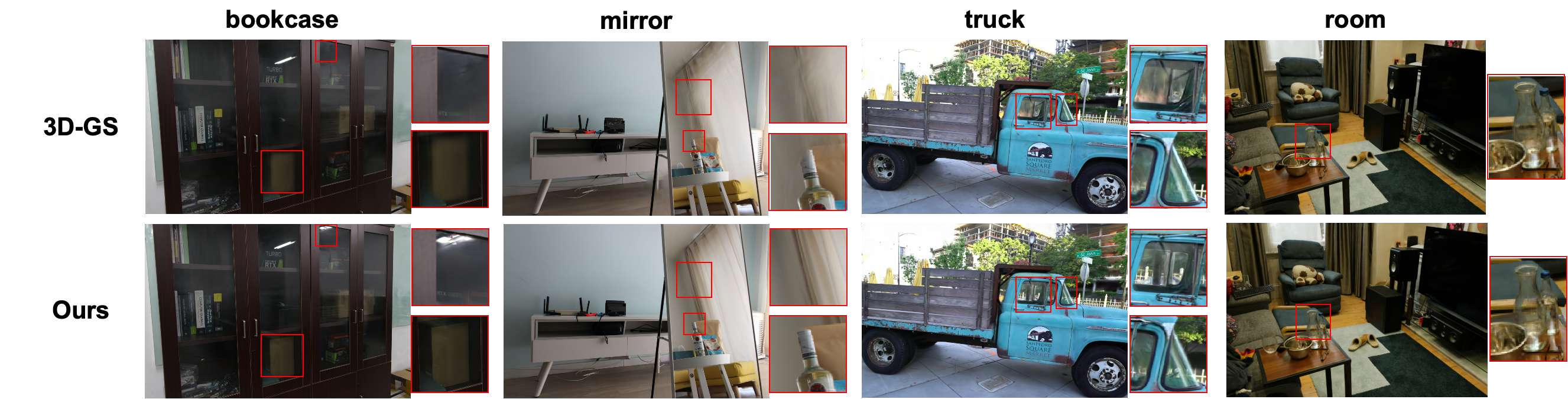}
    \caption{\textbf{Detailed visual comparisons between 3D-GS and our method.} For a comprehensive comparison, we further show visual results of 3D-GS and our method with zoom-in details on RFFR scenes \textbf{bookcase} and \textbf{mirror}, and more general scenes like \textbf{room} from the Mip-NeRF360 dataset and \textbf{truck} from Tanks\&Temples.}
    \label{fig:detailed}
\end{figure*}
\begin{figure}[ht]
    \centering
    \includegraphics[width=\linewidth]{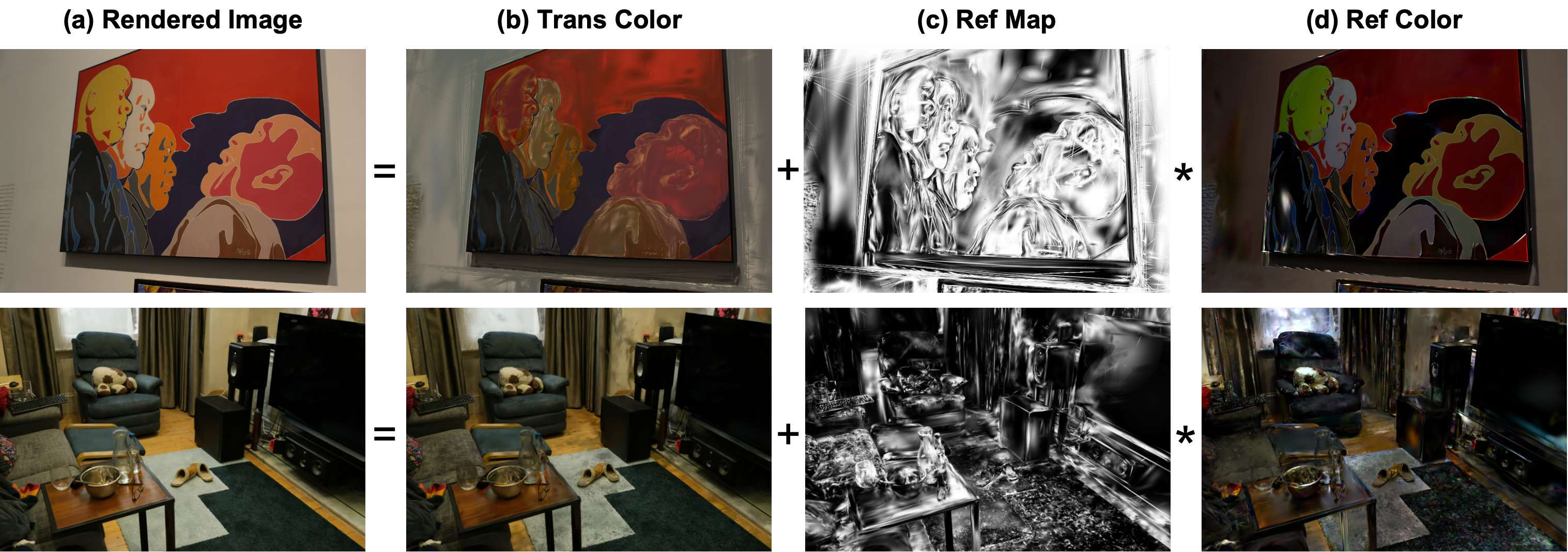}
    \caption{Illustration of scene disentanglement. The final rendered images (a) combine transmitted components (b) with the product of reflected components (c) and the reflection fraction map (d).}
    \label{fig:disentangle}
\end{figure}

\begin{figure}[ht]
    \centering
    \includegraphics[width=\linewidth]{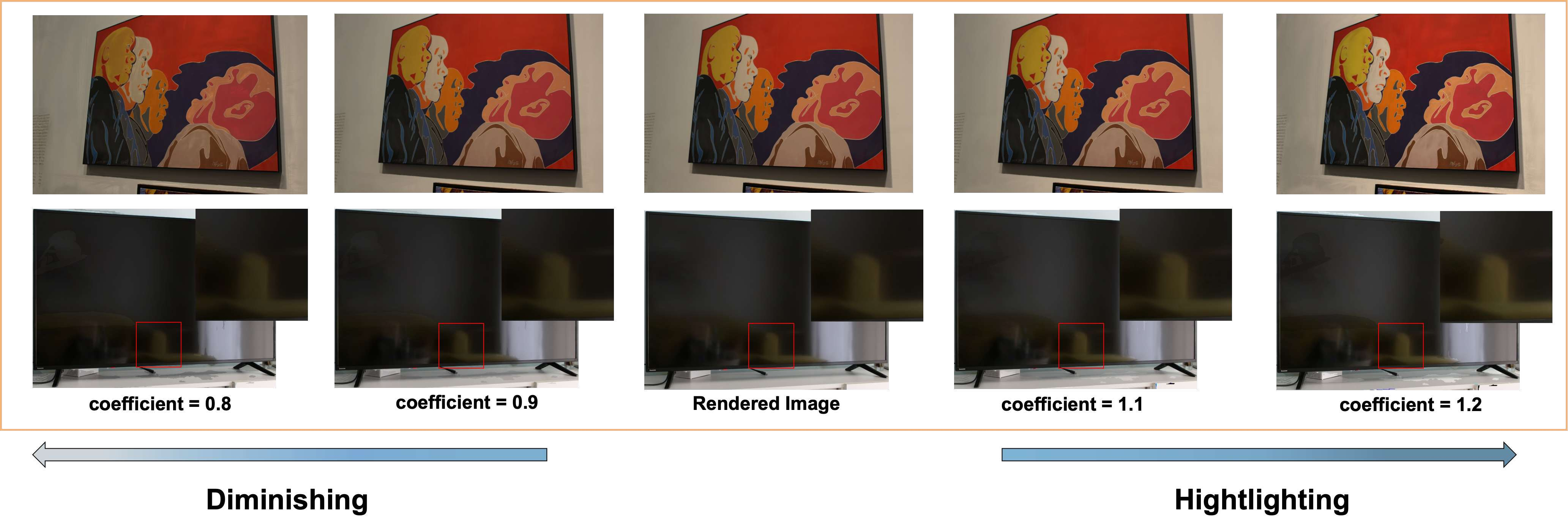}
    \caption{\textbf{Reflection Manipulation.} By adjusting the relighting coefficient (\textbf{0.8}, \textbf{0.9}, \textbf{1.0}, \textbf{1.1} and \textbf{1.2} from \textbf{left} to \textbf{right}) on the reflection map, we can arbitrarily augment or diminish the brightness of the reflected content. }
    \label{fig:manipulation}
\end{figure}

\begin{figure*}[t]
    \centering
    \includegraphics[width=\linewidth]{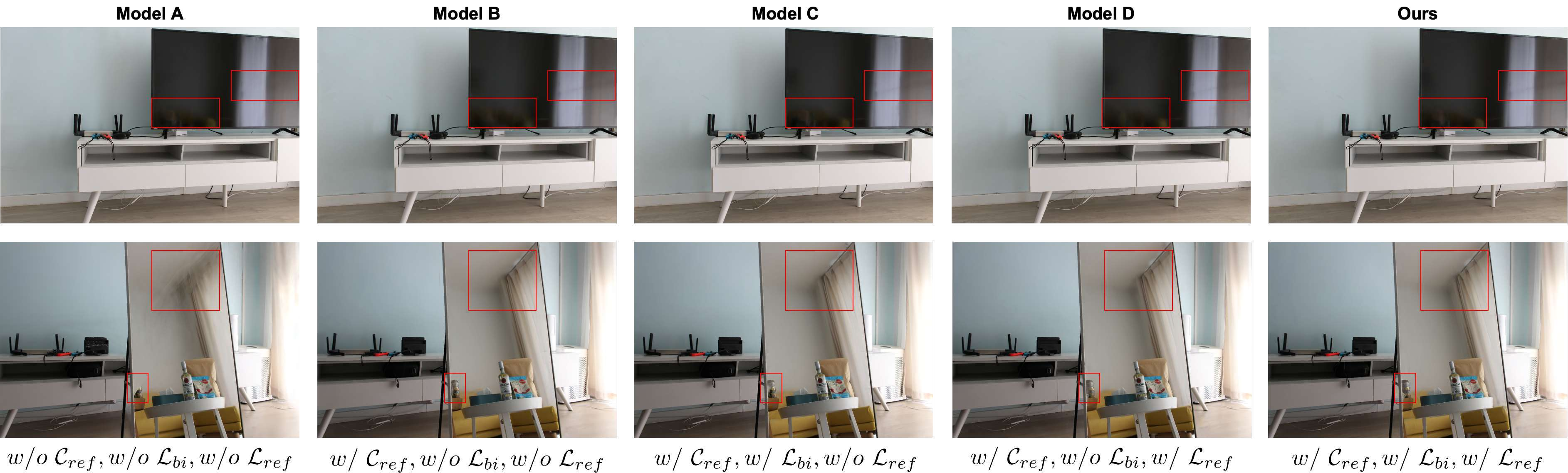}
    \caption{\textbf{Effectiveness of Model Design.} The combination of all three designs produces the best rendering results.}
    \label{fig:ablation}
\end{figure*}

\subsection{Bilateral Smoothness Priors}
%In our method, the transmitted component is supposed to focus on only the rendering of the realistic scene parts. 
The transmitted component is expected to focus solely on rendering realistic parts of the scene.
%\textcolor{red}{the logic needs to be improved.}
However, without the help of supervision from the reflection mask during the training process, there is a potential leakage of the reflected contents into the transmitted component and vice versa. This incomplete decomposition may cause an unsmoothed depth map and inconsistent rendering of color in the local region, thus further leading to unstable training.  

The depth constraint is employed in several methods~\citep{guo2022nerfren, chung2023depth, meng2024mirror, zhu2023fsgs} to circumvent the local inconsistency and improve the rendering quality.
The sole smoothing of the depth map works well in most cases, but it proves inadequate when rendering scenes containing reflections. This is due to its inability to identify and address the photometric inconsistency within the local region.
In scenes with reflections, the inaccurate estimation of depth for both the reflected content and the reflector significantly accounts for the unsatisfactory rendering results.
Thus, a new regularization scheme that jointly smooths the local geometry and the corresponding photometric consistency is more practical in such scenarios. In this sense, we propose a bilateral smoothness constraint on the rendered results to counteract the disturbance of the reflection and produce more reasonable local details. 
%Since commonly seen reflectors in the real world have rather smooth structures, e.g., mirrors, windows, and screens, this constraint assumes the reflectors in the scene to be planar surfaces rather than irregular surfaces with significant undulations or cusps. %
Specifically, this bilateral smoothness constraint takes the form of the bilateral filter, an edge-preserving and noise-reducing smoothing method for images. 
In brief, the bilateral filter combines the traditional domain filter that smooths the pixels with neighbors in the local region and the range filter that averages the pixels according to value similarity. 
Thus, such constraint is able to regularize each pixel with its 8 adjacent pixels considering both geometric closeness and photometric similarity. For the image $I_d$ under a random viewpoint $d$, the constraint is defined as follows:
\begin{equation}
    \mathcal{L}_{bi} = {\sum_{p_i{\in}\boldsymbol{I_d}}}{\sum_{p_j{\in}\boldsymbol{\mathcal{N}_i}}} f(p_i, p_j)||\widehat{D}_{p_i} - \widehat{D}_{p_j}||_1
    \label{eq:bilateral smooth}
\end{equation}
\begin{equation}
    f(p_i, p_j) = exp(\dfrac{-||\widehat{C}_{p_i} - \widehat{C}_{p_j}||_1}{\gamma})
\end{equation}
where $\mathcal{N}_i$ refers to the neighbor set of $p_i$, function $f(p_i, p_j)$ projects the photometric difference between pixels into weights, $\gamma$ is the scale factor, $\widehat{D}_p$ is the estimated depth of the rendered pixel and the $\widehat{C}_p$ is the color of the rendered pixel. 
Weights $f(p_i, p_j)$ are regularized along with the local depth map while also used to relax the depth constraint on the edge areas that present significant depth changes. 
\citep{guo2022nerfren} leverages a similar constraint called depth smoothness utilizing the idea of bilateral filter. However, it utilizes the photometric differences between the ground truth pixels as the weights, thus neglecting the regularization on the rendered colors that are intrinsically related to the depth information. 

Besides, since the rendering process is disentangled into two components, we also regularize the training of the reflected component with a reflection map smoothness:
\begin{equation}
    \mathcal{L}_{ref} = {\sum_{p_i{\in}\boldsymbol{I_d}}}{\sum_{p_j{\in}\boldsymbol{\mathcal{N}_i}}} ||\widehat{W}_{p_i} - \widehat{W}_{p_j}||_1
    \label{eq:ref map smooth}
\end{equation}
where $\widehat{W}_p$ is the estimated reflection weight of the pixel. 
With the above two smoothness constraints, we are able to alleviate the mutual perturbations between the two parts and present a more complete decomposition. 

\subsection{Optimization}
We optimize the attributes of Gaussians with both the photometric loss and the aforementioned constraints. The photometric loss $\mathcal{L}_{rgb}$ is the combination of the $\mathcal{L}_1$ and $\mathcal{L}_{D-SSIM}$:
\begin{equation}
    \mathcal{L}_{rgb} = \lambda \mathcal{L}_1(C, \widehat{C}) + (1 - \lambda) \mathcal{L}_{D-SSIM}(C, \widehat{C})
\end{equation}
where $\lambda$ is the balance coefficient, $C$ is the ground truth image and $\widehat{C}$ is the rendered image. $\mathcal{L}_1$ calculates the absolute error between inputs while $\mathcal{L}_{D-SSIM}$ refers to the differentiable structural similarity index measure. 

We additionally use a $\mathcal{L}_1$ loss in the first few iterations of training to force an alignment between the ground truth image and the transmitted image for stable training and fast convergence:
\begin{equation}
    \mathcal{L}_{init} = \mathcal{L}_1(C, \widehat{C}_{trans})
\end{equation}

The overall loss used in our RefGaussian can be formulated as follows:
\begin{equation}
    \mathcal{L}_{overall} = \mathcal{L}_{rgb} + \lambda_{init} \mathcal{L}_{init} + \lambda_{bi} \mathcal{L}_{bi} + \lambda_{ref} \mathcal{L}_{ref}
\end{equation}
where $\lambda_{init}$, $\lambda_{bi}$ and $\lambda_{ref}$ are the coefficients of each loss term.

\section{Experiments}
\label{sec:Experiments}
In this section, we demonstrate the superior performance of our RefGaussian with a holistic set of experiments. 
We first present both the quantitative and qualitative comparisons with four competitive methods in Sec.~\ref{sec:comparisons}. Then, we show the disentanglement results of the scenes in Sec.~\ref{sec:disentanglement}.
We also show in Sec.~\ref{sec:reflection manipulation} that, with the disentangled design of RefGaussian, the reflection manipulation is achievable at a pixel level. Finally in Sec.~\ref{sec:alation}, we conduct several ablation studies on the model design and choices of loss functions. 

\textbf{Baselines.}
We compare our RefGaussian with NeRF~\citep{reiser2021kilonerf}, NeRFReN~\citep{guo2022nerfren}, NeRF-D~\citep{guo2022nerfren} and 3D-GS~\citep{kerbl20233d}. 
Note that NeRFReN utilizes manually annotated masks to guide the decomposition of the \textit{mirror} and \textit{tv} scenarios, thus achieving impressive decomposition results. 
For a fair comparison, we remove the mask supervision and reproduce the experiments with NeRFReN under our train-test-split setting. 
Commonly used image quality metrics, i.e. PSNR, SSIM~\citep{wang2003multiscale} and LPIPS~\citep{zhang2018unreasonable}, are applied to evaluate all methods.

\textbf{Datasets.}
We conduct the main experiment on the RFFR
(Real Forward-Facing with Reflections) dataset~\citep{guo2022nerfren}, which is specifically designed for the view synthesis task under strong reflections. This dataset contains 6 captured forward-facing scenes with strong reflection, including \textit{art1}, \textit{art2}, \textit{art3}, \textit{bookcase}, \textit{tv}, and \textit{mirror}. Besides, to further validate the general performance of our method on the random scenes, we also test it on \textit{room} from the Mip-NeRF360 dataset~\citep{barron2022mip} and \textit{truck} from Tanks \& Temples~\citep{knapitsch2017tanks}. 
For each scene, we randomly shuffle the data and select one out of every eight images as the test set and gather the rest as the training set. 

\textbf{Implementation Details.}
For training and testing, we downsample the input images to a fixed $1296 \times 864$ size for all scenes. 
We train the 3D-GS and RefGaussian for 30,000 iterations with loss weights $\lambda_1$, $\lambda_{D-SSIM}$, $\lambda_{init}$, $\lambda_{bi}$ and $\lambda_{ref}$ set to 0.8, 0.2, 0.1, 0.0001 and 0.0001, respectively. 
NeRF-based methods are trained for 40 epoches with a batch size of 128 under the official code. 
All the experiments are conducted on NVIDIA GeForce RTX 3090 GPU.

\subsection{Comparisons}
\label{sec:comparisons}
\textbf{Quantitative Results.} As presented in Table~\ref{tab:main}, our RefGaussian achieves the best overall performance in terms of PSNR and SSIM. In particular, RefGaussian outperforms all methods in scenes with macroscopic reflections, i.e. \textit{bookcase}, \textit{tv} and \textit{mirror} scenarios as shown in Figure~\ref{fig:main}. For \textit{art} scenes in which reflections are less obvious, our method is on par and mostly surpasses other 3D-GS and NeRF-based methods. As to the LPIPS metric, RefGaussisn reaches the lowest perceptual loss in \textit{bookcase} against other methods. Although RefGaussian falls behind NeRF-based methods in the rest of the evaluation scenarios by a small margin, it still showcases an average performance gain over 3D-GS.

\noindent\textbf{Qualitative Analysis.}
We provide a holistic visual comparison between all five tested methods in Figure~\ref{fig:main}. Our method exhibits comparable performance with NeRF-based methods in scenes with semi-reflections and outperforms all other methods in scenes with specular reflections, e.g., \textit{mirror} and \textit{tv}. The result is remarkable since our method achieves both effectiveness and efficiency without mask supervision.
In Figure~\ref{fig:detailed}, we show a detailed comparison between RefGaussian and 3D-GS with zoom-in details. Evidently, our method demonstrates outstanding rendering quality in reflected areas. Moreover, in more general cases like \textit{truck} and \textit{room}, which contain only a small fraction of reflections, our method is still able to properly reconstruct the easily neglected reflections. 

\noindent\textbf{Rendering Speed. }
We also compare the rendering speed of the models in Table~\ref{tab:main} under the metric FPS (Frames Per Second) and present the results in Table~\ref{tab:fps}. 
Since we implement a synchronous rendering of both the transmitted and the reflected components, our method achieves an average of 81.856 FPS on a single NVIDIA GeForce RTX 3090 GPU, which is moderately lower than the original 3D-GS. 
However, compared to NeRF-based methods like NeRFReN and NeRF-D, RefGaussian achieves a significant improvement in rendering speed while maintaining comparable rendering quality.
These results demonstrate that our RefGaussian showcases an optimal trade-off between the model performance and the computational efficiency. 

\begin{table}[t]
\centering
\setlength{\tabcolsep}{5pt}
\caption{Ablation Studies on Model Design.}
\label{tab:albation on model design}
{
\begin{tabular}{l|ccc|ccc}
\bottomrule[1.5pt]
  Model  & Reflection\_SH & Bi\_Loss  & Ref\_Loss  & PSNR $\uparrow$ & SSIM $\uparrow$  & LPIPS $\downarrow$ \\ \hline
A  &     &     &           &  33.978     &  0.956     & 0.187  \\  
B  &    \multicolumn{1}{c}{\checkmark}  &     &           &   34.172
    &      0.954 & 0.204 \\
C &   \multicolumn{1}{c}{\checkmark} &   \multicolumn{1}{c}{\checkmark}  &      &   34.429
      &   0.956  &    0.180\\
D &  \multicolumn{1}{c}{\checkmark}  &     &   \multicolumn{1}{c|}{\checkmark}   &    34.667
     &  0.957 &   0.180 \\ \hline
Ours &   \multicolumn{1}{c}{\checkmark} &   \multicolumn{1}{c}{\checkmark}  &   \multicolumn{1}{c|}{\checkmark}   &         \textbf{34.739}  &     \textbf{0.957}    &     	\textbf{0.177}\\
\toprule[1.5pt]
\end{tabular}
% \vspace{-0.6cm}
}
\end{table}

\begin{table}[t]
% \footnotesize
\centering
\caption{Ablation Studies on the Weights of Bilateral Smoothness.}
\label{tab:bilateral weight}
\begin{tabular}{lcccccc}
\toprule[1.5pt]
     \multicolumn{1}{c}{$\lambda_{bi}$}  & 0.01 & 0.001 & 0.0001 & 0.00001 \\ \midrule[1pt]
PSNR $\uparrow$   &    31.465 &  32.821 &  \textbf{34.859}
   &  34.250 \\
SSIM $\uparrow$    &  0.928  &  0.948  &   \textbf{0.957}  &  0.955 \\
LPIPS $\downarrow$    &   0.257    &  0.197 &  \textbf{0.177}  & 0.180  \\ \bottomrule[1.5pt]
\end{tabular}
% \vspace{-0.6cm}
\end{table}

\begin{table}[ht]
% \footnotesize
\centering
\caption{Ablation Studies on Weights of Reflection Map Smoothness.}
\label{tab:reflection weight}
\begin{tabular}{lcccccc}
\toprule[1.5pt]
     \multicolumn{1}{c}{$\lambda_{ref}$}  & 0.01 & 0.001 & 0.0001 & 0.00001 \\ \midrule[1pt]
PSNR $\uparrow$ &    33.858    &   34.568 &   \textbf{34.859}  &  34.501 \\
SSIM $\uparrow$    &   0.955 &  0.956  &  \textbf{0.957}   &  0.955 \\
LPIPS $\downarrow$    &   0.181    &  0.180 &  \textbf{0.177}  &  0.181 \\ \bottomrule[1.5pt]
\end{tabular}
% \vspace{-0.6cm}
\end{table}

\subsection{Reflection Disentanglement}
\label{sec:disentanglement}
To gain deep insight into the effectiveness of our RefGaussian, we further conducted experiments on reflection disentanglement for view synthesis.
As shown in Figure~\ref{fig:disentangle}, the transmitted components, reflection color, and reflection map are well disentangled.
The final render images combine transmitted components and reflected components multiplied by the reflection fraction map using Eq.~\ref{eq.6}.
In contrast to NeRFReN, which relies on manually annotated reflection masks as supplementary information, all of our experiments on reflection disentanglement are conducted using RefGaussian, without any additional assumptions.
The results presented in Figure~\ref{fig:disentangle} demonstrate that RefGaussian effectively accomplishes transmission-reflection decomposition, aligning with human perception while delivering high-quality view synthesis results. 
The reflection fraction map acquired through learning provides a rough indication of the surface material, with larger values (white) representing reflectors and smaller values (black) representing non-reflectors. 
It is important to note that our primary focus is on modeling the stable virtual image through the reflected field while leaving the low-frequency highlights to the view dependency within the transmitted field. 
Accurately estimating the position of transparent surfaces, such as glass, poses challenges due to the absence of easily traceable image features. 
In cases like up-row in Figure~\ref{fig:disentangle}, RefGaussian treats the opaque surface behind as the reflective surface, which is not strictly accurate, but it does not significantly impact the quality of view synthesis due to image-space composition.
In more complex scenarios like bottom-row in Figure~\ref{fig:disentangle}, RefGaussian also performs well without any plane or mask assumptions, demonstrating the versatility of RefGaussian.

\subsection{Reflection Manipulation}
\label{sec:reflection manipulation}

Unlike most methods that realize scene editing with the help of the reflection masks~\citep{guo2022nerfren}, RefGaussian conducts reflection manipulation in a more simple and straightforward manner by multiplying the reflection fraction map by a relighting coefficient that controls the intensity. 
Such manipulation is achieved by simulating the principles of light reflection and absorption in the real physical world instead of directly operating on the image pixels. As is shown in Figure~\ref{fig:manipulation}, we can adjust the brightness of the reflections at the arbitrary scale by controlling the coefficient. Both the highlighting and the diminishing process are smooth and realistic as if the the transmission of the lights changes as the coefficient scales. 
% \textbf{Reflection Reduction}

% \textbf{Reflection Increment}

\subsection{Ablation Studies}
\label{sec:alation}

All the ablation studies are conducted on the RFFR dataset to validate the model design in our method and the effect of hyperparameters. All the metrics are reported in the form of averages across the dataset. 

\textbf{Effectiveness of Model Design.}
To validate our model design, we trained four extra models to test the effectiveness of our disentangled scheme and carefully designed constraints. 
Table~\ref{tab:albation on model design} shows the detailed setting of each model and the corresponding quantitative results. The sole employment of reflection disentanglement (Model B) improves the model performance only in terms of PSNR compared to the base model (Model A), but reports an overall enhancement when combined with either of the two smoothness constraints (Model C and Model D). Evidently, our method integrates all three components and demonstrates the best performance across all three metrics. 

The visual results are depicted in Figure~\ref{fig:ablation}. Compared to the base Model A, all three models (Model B, C, and D) with the reflection disentanglement design are able to accurately reconstruct the reflected curtains in scene \textit{mirror} and the reflected railings in scene \textit{tv} (see details in the red box). 
However, Model C with the bilateral smoothness prior fails to explicitly synthesize the beverage can in scene \textit{mirror} while Model D with the reflection map smoothness only generates a blurry rendering of the reflection of the sofa in scene \textit{tv}. By contrast, RefGaussian achieves high-fidelity rendering results in both scenes.   

\textbf{Ablation Studies on the Weight of Bilateral Smoothness.}
Table~\ref{tab:bilateral weight}.
We conduct four sets of experiments with $\lambda_{bi}=$ 0.01, 0.001, 0.0001 and 0.00001 to decide the optimal parameter for the weight of the bilateral smoothness prior. Table~\ref{tab:bilateral weight} shows that the model performs best when $\lambda_{bi} = 0.0001$. 

\textbf{Ablation Studies on the Weight of Reflection Map Smoothness.}
Table~\ref{tab:reflection weight}.
Similarly, we also run four sets of experiments to decide the weight of the reflection map smoothness constraint. The weight $\lambda_{ref}$ is set to 0.01, 0.001, 0.0001, and 0.00001, respectively. Table~\ref{tab:reflection weight} shows that setting $\lambda_{ref}$ to 0.0001 produces the best results. 
Notably, the weight of the bilateral smoothness prior is fixed to 0.0001 during the ablation experiments on the weight of reflection map smoothness, and vice versa.

\section{Limitations and Future Works}
\label{sec:limitation}
Since commonly seen reflectors in the real world have rather smooth structures, e.g., mirrors, windows, and screens, the proposed bilateral smoothness constraint and reflection map constraint assume the reflectors in the scene to be planar surfaces rather than irregular surfaces with significant undulations or cusps. Therefore, the applicability of our method in more diverse scenarios, e.g., scenes with curved reflectors, is currently unknown. Besides, the reflection manipulation in our RefGaussian is achieved at the pixel level through filtering the reflection map, thus producing only coarse and global enhancement. More flexible and fine-grained manipulation is a focus of our future works.  

\section{Conclusions}
\label{sec:conclusion}
In this work, we propose RefGaussian, a novel 3D-GS-based framework tailored for the realistic rendering of scenes with strong reflections. Creatively, our RefGaussian decomposes the scene into the transmitted component and the reflected component and fuses the renderings of two components to generate the final results. Accordingly, Gaussian representation is extended with three additional parameters, namely, reflection-SH, reflection opacity, and reflection confidence to support the disentangled scheme. 
Further, we introduce the bilateral smoothness and the reflection map smoothness constraints to remove the effect of mutual interference between two components and ensure proper decomposition and improved performance.
Our method outperforms NeRF-based methods and original 3D-GS by a large margin on scenes with strong reflections and also shows comparable results on more general scenes. 
Finally, the capability of the reflection manipulation demonstrates a promising potential of our method in broader applications.

\bibliographystyle{unsrtnat}
\bibliography{ref}  %%% Uncomment this line and comment out the ``thebibliography'' section below to use the external .bib file (using bibtex) .





% \end{thebibliography}

\end{document}